\documentclass[10pt,twocolumn,letterpaper]{article}

\usepackage[pagenumbers]{cvpr} 

\definecolor{cvprblue}{rgb}{0.21,0.49,0.74}
\usepackage[pagebackref,breaklinks,colorlinks,allcolors=cvprblue]{hyperref}
\usepackage{wrapfig}
\usepackage{algorithm, algorithmic}
\usepackage[accsupp]{axessibility}

\definecolor{RS}{RGB}{0,0,128}

\def\paperID{11965} 
\def\confName{CVPR}
\def\confYear{2025}

\title{Efficient Fine-Tuning and Concept Suppression for Pruned Diffusion Models}
\author{
Reza Shirkavand$^{1}$,
Peiran Yu$^{2}$,
Shangqian Gao$^{3}$,
Gowthami Somepalli$^{1}$,
Tom Goldstein$^{1}$,
Heng Huang$^{1}$\\
\and
$^{1}$ University of Maryland, College Park\\
{\tt\small \{rezashkv,gowthami,tomg,heng\}@cs.umd.edu}
\and
$^{2}$ University of Texas, Arlington\\
{\tt\small peiran.yu@uta.edu}
\and
$^{3}$ Florida State University\\
{\tt\small sgao@cs.fsu.edu}
}
\begin{document}
\maketitle

\begin{abstract}
Recent advances in diffusion generative models have yielded remarkable progress. While the quality of generated content continues to improve, these models have grown considerably in size and complexity. This increasing computational burden poses significant challenges, particularly in resource-constrained deployment scenarios such as mobile devices. The combination of model pruning and knowledge distillation has emerged as a promising solution to reduce computational demands while preserving generation quality. However, this technique inadvertently propagates undesirable behaviors, including the generation of copyrighted content and unsafe concepts, even when such instances are absent from the fine-tuning dataset.
In this paper, we propose a novel bilevel optimization framework for pruned diffusion models that consolidates the fine-tuning and unlearning processes into a unified phase. Our approach maintains the principal advantages of distillation—namely, efficient convergence and style transfer capabilities—while selectively suppressing the generation of unwanted content. This plug-in framework is compatible with various pruning and concept unlearning methods, facilitating efficient, safe deployment of diffusion models in controlled environments. Code is available \href{https://github.com/rezashkv/unlearn-ft}{here}.

\end{abstract}    
\section{Introduction}
\label{sec:intro}

The rapid advancement in generative models, particularly diffusion models~\cite{diff-sohl,score-match-song,ddpm-ho,score-sde-song}, has led to the development of powerful tools capable of generating high-quality synthetic images~\cite{ldm-stable,imagen,sdxl}. However, the parameter-heavy architectures and significant memory requirements of these models often make their deployment on smaller GPU clouds and edge devices challenging. To address this resource-intensive nature, various approaches have been proposed~\cite{mobile-diff,BKSDM,ldm-slim,snapfusion,q-diff}, with model pruning~\cite{pruning-survey} being a prominent technique aimed at reducing the computational demands of diffusion models to improve efficiency~\cite{spdm,laptop-diff,aptp}. While pruning can significantly alleviate computational costs, retraining is typically needed to restore the pruned model's performance. Previous methods~\cite{BKSDM,laptop-diff,aptp} have utilized knowledge distillation~\cite{knowledge-distill,feature-dit} to fine-tune pruned models effectively.

\begin{figure}
    \centering
    \includegraphics[width=\linewidth]{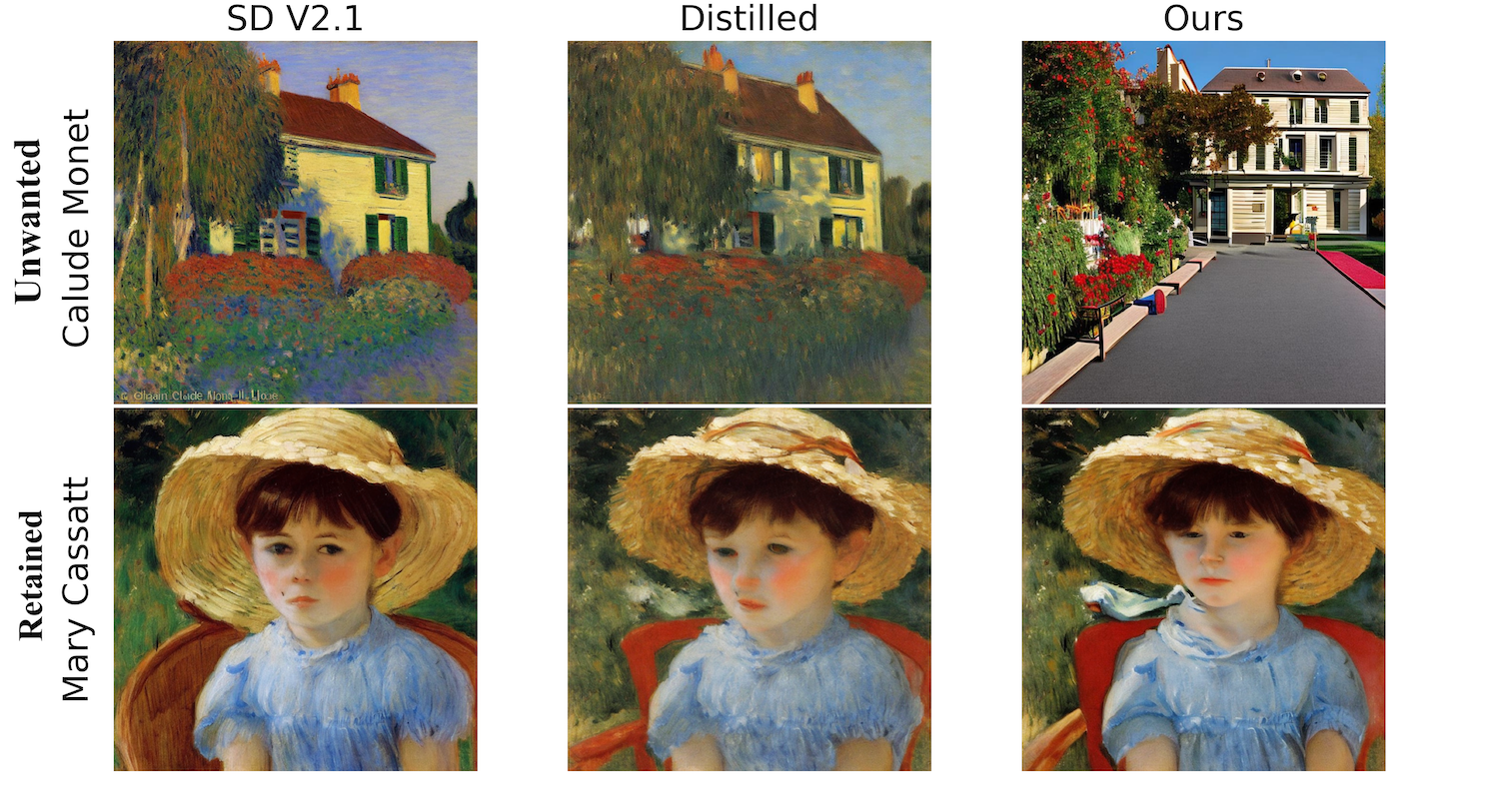}
    \caption{Comparison of images generated in the styles of Claude Monet (top row) and Mary Cassatt (bottom row), both impressionist artists, using Stable Diffusion, a pruned model finetuned using standard knowledge distillation, and our proposed controlled fine-tuning method. While both the Stable Diffusion and distilled models generate images in Monet's style, our method effectively unlearns Monet’s Impressionist style while preserving Cassatt’s distinct Impressionist features. This indicates our approach’s advantage in selectively suppressing specific styles while maintaining high fidelity to the other unremoved concepts and features. }
    \label{fig:main-teaser}
    \vspace{-10pt}
\end{figure}

\begin{figure*}
    \centering
    \includegraphics[width=\textwidth]{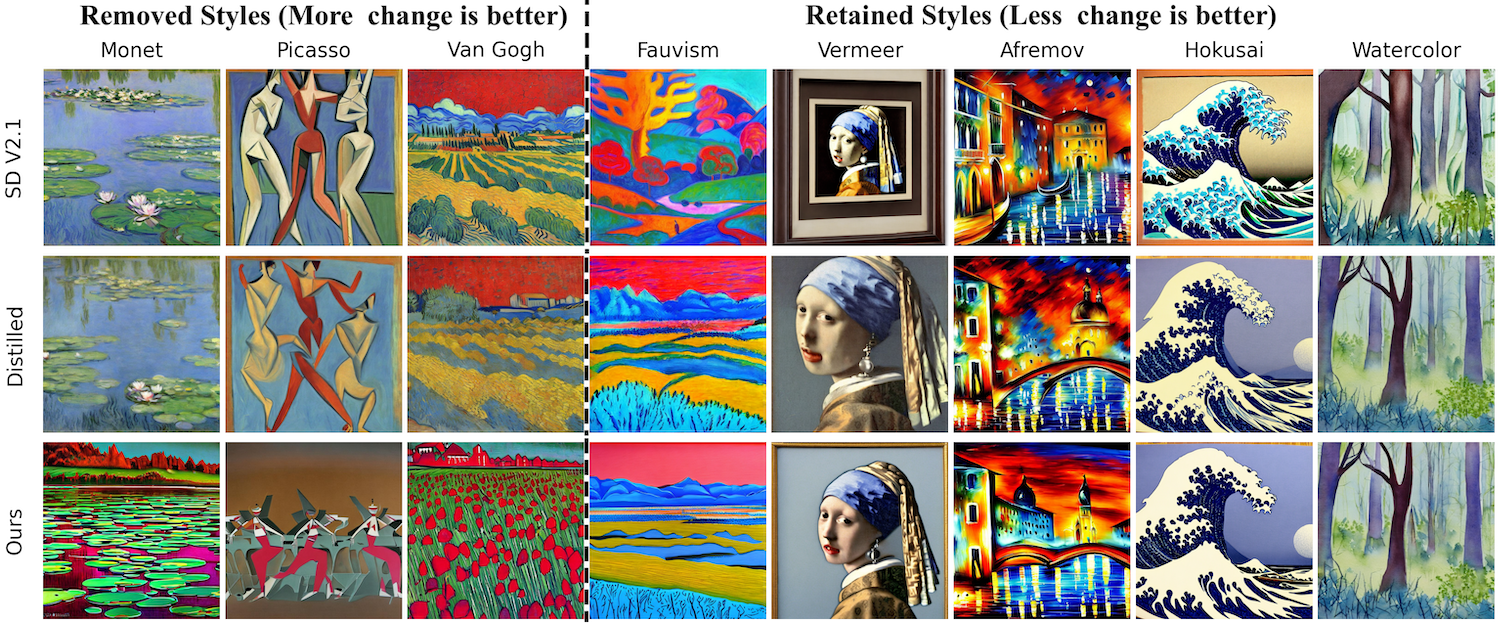}
    \caption{Comparison of generative quality and style adherence: \textbf{Row 1:} The original Stable Diffusion 2.1 model. 
    \textbf{Row 2:} A pruned version fine-tuned with 20,000 iterations of combined DDPM and distillation loss. \textbf{Row 3:} A pruned version fine-tuned with 20,000 iterations of our proposed bilevel fine-tuning approach, removing styles of Van Gogh, Monet, and Picasso. Our bilevel method is successful in retaining generative quality and style diversity while suppressing undesirable concepts. See the ~\cref{sec:app-exp_sub:prompts} for prompts used.}
    \label{fig:teaser}
    \vspace{-10pt}
\end{figure*}

While distillation enhances convergence speed and preserves the expressive power of the original model, it introduces inherent risks. Diffusion models are known to generate copyrighted or inappropriate (e.g., NSFW) content~\cite{red-teaming-sd-nsfw,diff-copy}, prompting ongoing efforts to filter out such content~\cite{erasing-dm,ablating-concepts,safe-ldm,genererate-or-not,pruning-for-erasing,editing-explicit-content,zhang2024defensive-adv-unlearning,huang2023receler} without the significant expense of retraining on a filtered dataset. It might seem intuitive that removing unsafe concepts from the fine-tuning dataset would naturally filter out such content. However, this assumption does not necessarily hold. Even when such content is excluded from the fine-tuning dataset, unsafe concepts still persist in a pruned diffusion model, particularly when distillation is used during the fine-tuning process (See \cref{fig:main-teaser,fig:teaser}). This poses a critical challenge for deploying diffusion models in controlled environments, where generating these types of outputs is unacceptable.

A naive approach to tackle this issue is to use distillation to recover the generative capabilities of the base model first, followed by a concept unlearning method to remove undesirable content. However, this approach is inefficient and, as we hypothesize and empirically demonstrate, suboptimal in practice. It often leads to ineffective concept removal and degraded generation quality due to the interdependent optimization requirements for both retraining and unlearning processes. Specifically, the parameters optimal for retraining to restore model performance are not necessarily the best initialization point for effective concept unlearning. Conversely, the parameters chosen for unlearning can impact the model's ability to regain its generative capabilities. This mutual dependency creates a circular optimization problem where the success of one step depends on the outcomes of the other, resulting in suboptimal trade-offs between preserving generative quality and removing unwanted concepts.

To effectively tackle this circular dependency challenge during the fine-tuning phase, we propose a novel bilevel optimization framework for fine-tuning pruned diffusion models. Our approach retains the benefits of distillation—rapid convergence and effective style transfer—while selectively suppressing unwanted generative behaviors from the base model. As we will show, it is significantly superior to the two-stage approach (first fine-tune then forget). The lower-level optimization performs standard distillation and diffusion loss minimization on the fine-tuning dataset to restore the model's generative capabilities. Meanwhile, the upper-level directs the model away from generating unwanted concepts.
Our method is a plug-in technique that can be integrated with various pruning methods for fine-tuning. It can also incorporate any concept unlearning method in the upper level optimization step.

To summarize our contributions:
\begin{itemize}
    \item We introduce a novel bilevel optimization framework for fine-tuning pruned diffusion models, effectively solving the interdependence problem between restoring generative quality and removing undesirable content. By integrating these tasks, our method avoids the inefficiencies and suboptimal results of sequential approaches, where the circular dependency between retraining and unlearning results in degraded model performance.

    \item 
    Our framework is designed to be adaptable, allowing the fine-tuning of the result of any pruning method. It can also incorporate any concept unlearning technique in the upper-level optimization. This flexibility broadens the applicability of our method to a wide range of resource-constrained settings and customization needs.

    \item 
    Through extensive evaluations of artist style and NSFW content removal tasks, we demonstrate that our bilevel method significantly outperforms the two-stage approach. Our results show superior concept suppression while retaining high generation quality, underscoring the effectiveness of our method in real-world controlled deployment environments.
    
\end{itemize}

\section{Related Work}
\label{sec:rel-work}

\subsection{Efficient Diffusion Architectures}
\label{sec:rel-work_sub:pruning}
Numerous studies have targeted efficient architecture design for diffusion models~\cite{mobile-diff,ldm-slim,snapfusion}. For instance, MobileDiffusion~\cite{mobile-diff} applies empirical adjustments based on performance metrics from MS-COCO~\cite{mscoco}, while SnapFusion~\cite{snapfusion} focuses on finding optimal architectures specifically for text-to-image (T2I) models. Pruning-based approaches~\cite{ganjdanesh2024mixture} attempt to eliminate non-essential layers or blocks of a pre-trained diffusion model. SPDM~\cite{spdm} evaluates the importance of weights via Taylor expansion, removing lower-value weights, while BK-SDM~\cite{BKSDM} streamlines the U-Net by eliminating blocks that minimally impact generation quality. APTP~\cite{aptp} employs a dynamic, prompt-based pruning strategy for T2I models, adjusting resources according to prompt complexity in a Mixture of Experts setting. Some pruning-based approaches utilize distillation techniques—both knowledge distillation~\cite{knowledge-distill} and feature distillation~\cite{feature-dit}—to retrain the pruned model~\cite{BKSDM,aptp,progressive-distill-sdxl,koala,laptop-diff}, restoring its generative abilities. 

To the best of our knowledge, previous works have neither thoroughly quantified the benefits of distillation in term of convergence speed of the pruned model, nor examined its potential to transfer undesirable properties from the base model to the pruned model.

\subsection{Concept Editing and Unlearning in Diffusion Models}
\label{sec:rel-work_sub:erasing}
Most concept editing techniques in diffusion models work by aligning model outputs with a reference prompt that retains desired features. For example, removing “Van Gogh” style might involve pairing “a painting of a woman in the style of Van Gogh” with “a painting of a woman.”. Various strategies are used for this alignment, including minimizing certain metrics between the denoised outputs of target and reference prompts~\cite{ablating-concepts}, leveraging score-based unsupervised training data~\cite{erasing-dm}, or modifying cross-attention weights~\cite{forget-me-not}. Methods like UCE~\cite{uce} achieve concept removal by adjusting token embeddings directly through parameter changes in the U-Net's attention module.  

Diffusion model unlearning (MU) methods~\cite{erasediff-dual,separable-concept-ersure,geom-erasing,salun,continual-learning-forgetting,zhang2024defensive-adv-unlearning}, by contrast, rely on a “forgetting” dataset to remove specific information. These methods employ optimization approaches like dual problem formulations~\cite{erasediff-dual}, generative replay for reinforcing retention~\cite{continual-learning-forgetting}, and saliency-based fine-tuning masks~\cite{salun}. Unlike these approaches, ConceptPrune~\cite{concept-prune} offers a training-free pruning method to remove regions responsible for undesired outputs.

While these methods focus primarily on concept editing or unlearning, sometimes introducing sparsity to ensure the model “forgets,” they do not emphasize achieving high levels of sparsity. In contrast, our work is focused on efficiently fine-tuning a pruned model that has been reduced to a target sparsity level. 
It results in quick and effective adaptation while also suppressing undesirable content.
\section{Preliminary}
\label{sec:prelim}

\subsection{Diffusion Models}
\label{sec:prelim_sub:diffusion}
Given a training dataset $D$ with an underlying distribution $p_d$ and standard deviation $\sigma_d$, diffusion models generate samples by reversing a noise-adding process~\citep{diff-sohl, ddpm-ho, score-sde-song}. This process gradually introduces Gaussian noise to an initial sample $x_0$ such that $x_t = \alpha_{t} x_{0} + \sigma_{t} \epsilon_{t}$, where $\epsilon_{t} \sim \mathcal{N}(0, I)$ represents standard Gaussian noise. The level of perturbation increases with $t \in [0, T]$, where larger values of $t$ indicate higher noise levels. The parameters $\alpha_t$ and $\sigma_t$ are selected according to the diffusion model formulation. For instance, EDM~\cite{edm} sets $\alpha_{t} = 1$ and $\sigma_{t} = t$. 

These models are trained to minimize the following objective:
\begin{equation}\label{eq:denoising-obj}
    \mathcal{L}_{D}^{\text{Diff}}(\theta) = \mathbb{E}_{x_0, t, c, \epsilon}[w(t) \| \epsilon_{\theta}(x_t, t, c) - \epsilon \|^2],     
    \vspace{-5pt}
\end{equation}
where $w(t)$ is a weighting function and $c$ represents any concept or object the model is conditioned on.

\subsection{Model Pruning \& Distillation}
\label{sec:prelim_sub:pruning}

Given the parameters $\theta$ of a pre-trained model, the objective of model pruning is to identify a sparse parameter set, $\theta_{\text{pruned}}$, that closely preserves the model’s original performance~\cite{conv-pruning, pruning-survey}. This can be formalized as minimizing the loss variation due to pruning:
\begin{equation}\label{eq:pruning-obj}
\vspace{-5pt}
\min_{\theta_{\text{pruned}}} |L(\theta_{\text{pruned}}) - L(\theta)|, \quad \text{s.t.} \quad \|\theta_{\text{pruned}}\|_0 \leq R,
\end{equation}
where $L$ is a loss term ensuring the pruned model performs as closely as possible to the original. $\|.\|_0$ represents the $L_0$ norm, measuring the number of remaining non-zero parameters, and $R$ specifies the desired sparsity level. 

After pruning, the model usually loses its high quality generation capabilities and requires an additional fine-tuning stage. 
Previous approaches to retraining pruned diffusion models~\cite{BKSDM,aptp,laptop-diff}, sometimes leverage additional objectives such as output distillation and feature distillation~\cite{knowledge-distill,feature-dit} together with the original denoising objective (\cref{eq:denoising-obj}). These objectives encourage the pruned model (student) to match the behavior of the original model (teacher) across both output predictions and intermediate feature representations. Specifically, the distillation objectives are defined as:
\vspace{-10pt}
\begin{align}
    \mathcal{L}_{D}^{\text{Out-KD}} &= \mathbb{E}_{x_0,\epsilon,t} \| \epsilon_{T}(x_t, t,c) - \epsilon_{S}(x_t, t,c) \|^2, \label{eq:out-kd}\\
    \mathcal{L}_{D}^{\text{Feat-KD}} &= \sum_{i} \mathbb{E}_{x_0, \epsilon,t} \| \epsilon^i_{T}(x_t, t, c) - \epsilon ^i_{S}(x_t, t, c) \|^2,\label{eq:out-feat-kd}
\end{align}
where $\epsilon_S$ refers to the output of the student (pruned U-Net), and $\epsilon_T$ refers to the output of the teacher (original U-Net). Additionally, $\epsilon^i_T$ and $\epsilon^i_S$ represent feature maps at the $i$-th stage (\eg $i$-th block or $i$-th layer) in the teacher and student models, respectively.

\subsection{Concept Unlearning}
Following prior work~\cite{erasing-dm, uce, ablating-concepts}, we define concept unlearning for a pre-trained generative model $p_{\theta}(x,c)$ as the task of preventing the generation of a specific concept $c$. A natural question then arises: what should be generated in place of this omitted concept? Previous approaches often guide the model to generate samples conditioned on a related anchor concept, denoted by $c'$. The anchor concept $c'$ could be a similar concept to the target concept $c$—for example, Anchor: "Cat" vs. Target: "Grumpy Cat"~\cite{ablating-concepts}. Alternatively, $c'$ could represent a "null" concept, which encourages the model to revert to generating samples that resemble the unconditional outputs of the pre-trained model~\cite{edm}. The unlearning objective would then be
\begin{equation}\label{eq:concept_ablation}
    \min_{\theta_{CU}}\mathcal{L^{\text{CU}}}=D_{KL}(p_{\theta}(x|c') || p_{\theta_{CU}}(x|c)),
\end{equation}
As shown in~\cite{ablating-concepts}, in the case of diffusion models, the objective in \cref{eq:concept_ablation} can be reformulated as :
\begin{equation}\label{eq:ablation_distillation}
     \min_{\theta_{CU}}\mathbb{E}_{x_0,\epsilon,t,c,c'} \| \epsilon_{\theta}(x_t, t,c') - \epsilon_{\theta_{CU}}(x_t, t,c) \|^2,
\end{equation}
This formulation is closely related to \cref{eq:out-kd}. Indeed it effectively performs a knowledge distillation of the anchor concept $c'$ from the pre-trained model $\epsilon_{\theta}(.)$ into the model being modified for concept unlearning, $\epsilon_{\theta_{CU}}(.)$.
At the same time, we aim to preserve the generation capabilities of the pre-trained model on unrelated concepts, denoted by $\bar{c}$, so that:
\begin{equation}\label{eq:concept_preservation} 
    D_{KL}(p_{\theta}(x|\bar{c}) \parallel p_{\theta_{CU}}(x|\bar{c})) \approx 0,
\end{equation}
In existing work on concept removal in diffusion models, this preservation is typically achieved by initializing $\theta_{CU}$ with $\theta$ and assuming that the distribution over unrelated concepts remains unchanged throughout the concept removal process. 
\section{Method}
\label{sec:method}
Assume we apply a diffusion pruning method~\cite{spdm,BKSDM,aptp} to introduce sparsity in a pre-trained diffusion model. Our proposed fine-tuning approach can then be applied following any pruning technique. The objective is to jointly fine-tune the pruned diffusion model through distillation while removing undesirable properties of the base model.  

Formally, let $\epsilon_{\theta_{\text{pruned}}}$ represent the pruned diffusion model. Given a fine-tuning dataset $D_f$, the overall fine-tuning objective when using knowledge distillation (\cref{eq:out-kd}) and feature distillation (\cref{eq:out-feat-kd}) is given by:
\begin{equation}\label{eq:ft_all}
\min_{\theta_{\text{pruned}}} \mathcal{L^{\text{ft}}}:= \mathcal{L}^{\text{Diff}} + \lambda^{\text{OutKD}} \mathcal{L}^{\text{OutKD}} + \lambda^{\text{FeatKD}} \mathcal{L}^{\text{FeatKD}},
\vspace{-5pt}
\end{equation}
where we omit dependence on $\theta_{\text{pruned}}$ and $D_f$ for brevity and $\mathcal{L^{\text{ft}}}$ represents the total fine-tuning loss. $\lambda^{\text{OutKD}}$ and $\lambda^{\text{FeatKD}}$ are weighting coefficients for each distillation term in the weighted average. As we will show, incorporating distillation loss terms improves the convergence speed and generation quality of the pruned model (\cref{fig:distill-effect}). Assume this optimization yields $\hat{\theta}$, where $\hat{\theta} \in \text{argmin}_{\theta_{\text{pruned}}} \mathcal{L^{\text{ft}}}$.

Suppose we want to remove an undesirable property of the base model. As shown in \cref{fig:teaser}, the pruned model can still generate all the styles and concepts of the base model, even if these are absent in $D_f$—especially when distillation is used. To eliminate the pruned model’s ability to generate certain concepts $c$ (e.g., Van Gogh style or NSFW content), we employ the concept unlearning objectives in \cref{eq:concept_ablation} and \cref{eq:ablation_distillation}.

\begin{figure}
    \centering
    \includegraphics[width=\linewidth]{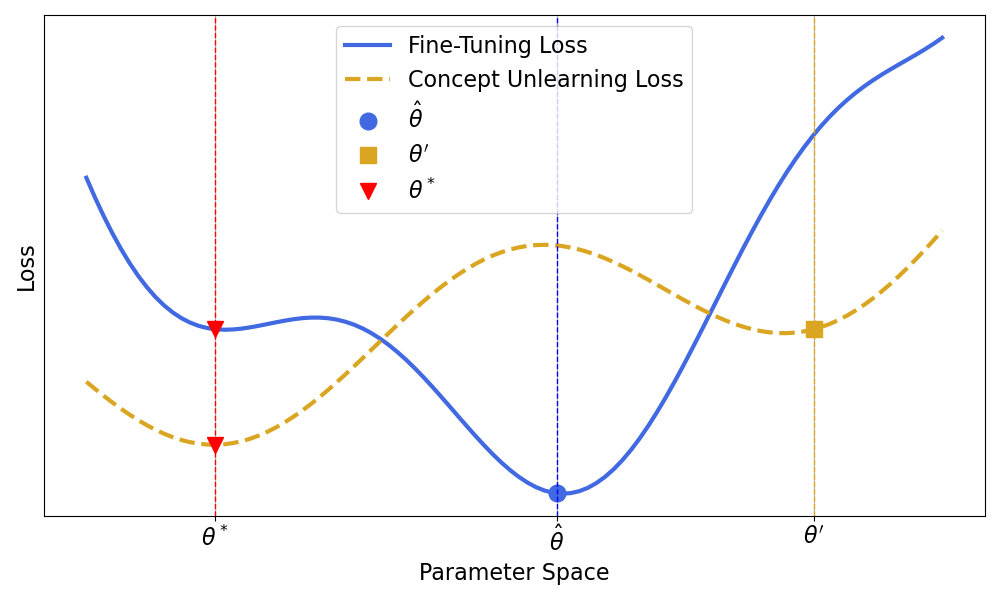}
    \caption{Why can a two-stage approach (fine-tuning followed by forgetting) be suboptimal? If fine-tuning yields $\hat{\theta}$, initializing the concept unlearning parameters with $\hat{\theta}$ and optimizing the concept unlearning loss (\cref{eq:concept_ablation}) results in $\theta'$, which is suboptimal for both fine-tuning and for concept unlearning. In contrast, our bilevel method~(\cref{eq:constrained}) produces the optimal solution $\theta^{*}$, achieving better performance for both fine-tuning and unlearning.}
    \label{fig:bilevel}
    \vspace{-10pt}
\end{figure}

A straightforward but naive approach to the overall pipeline might then involve two stages: (1) performing distillation on the pruned model to restore its generative capabilities, obtaining $\hat{\theta}$, and (2) applying a concept unlearning stage, initializing $\theta_{CU}$ with $\hat{\theta}$ as in previous approaches, yielding $\theta'$, where $\theta' \in \text{argmin}{\theta_{\text{CU}}} \mathcal{L^{\text{CU}}}$. However, as shown in \cref{fig:bilevel}, deviating from $\hat{\theta}$ may degrade generation quality, as $\theta'$ may not minimize \cref{eq:ft_all} optimally, as observed in prior work~\cite{erasing-dm,uce,concept-prune}. Consequently, this two-stage pipeline may lead to suboptimal results (see \cref{tab:artist_erasure}), potentially requiring iterative fine-tuning and unlearning steps to reach an improved solution.

To address this interdependency issue, we reformulate the fine-tuning process of the pruned model as a bilevel optimization problem, aiming to find parameters that optimize both generation quality and safety.
\begin{equation}\label{eq:constrained}
\begin{split}
    \min_{\theta_{\text{pruned}}}\mathbb{E}_{x_0,\epsilon,t,c,c'} & \| \epsilon_{\theta}(x_t, t,c') - \epsilon_{\theta_{\text{pruned}}}(x_t, t,c) \|^2,\\
     &s.t.\ \theta_{\text{pruned}}\in \text{argmin} \mathcal{L^{\text{ft}}}.
\end{split}
\end{equation}
Classical methods for solving bilevel problems, like the one in \cref{eq:constrained}, require calculating second-order information (see ~\cite{FrDoFrPo17, FrFrSaGrPo18, GhWa18} for examples). However, when fine-tuning diffusion models, these methods become highly costly due to significant computational and memory demands. Recently, new frameworks for bilevel optimization have been introduced~\cite{LuM24, ShenC23, LiuLYZZ24, KwonKWN23, LiuYWSL22} that rely only on first-order information, substantially reducing computational costs and making them highly suitable for fine-tuning diffusion models. We employ one of these methods to solve \cref{eq:constrained}.

More specifically, \cref{eq:constrained} is equivalent to the following constrained optimization problem: 
\begin{equation}\label{eq:ineq_constrained}
\begin{split}
    \min_{\theta_{\text{pruned}}}&\mathbb{E}_{x_0,\epsilon,t,c,c'}  \| \epsilon_{\theta}(x_t, t,c') - \epsilon_{\theta_{\text{pruned}}}(x_t, t,c) \|^2,\\
     s.t.\ & \mathcal{L^{\text{ft}}}(\theta_{\text{pruned}}) - \inf_{\vartheta}\mathcal{L^{\text{ft}}}(\vartheta)\le 0.
\end{split}
\vspace{-5pt}
\end{equation}
In ~\cref{eq:ineq_constrained} $\theta_{\text{pruned}}$ represents the parameters updated during the unlearning process, while $\vartheta$ denotes the parameters used for the fine-tuning task. These sets may be the same or different, depending on the specific choices made for unlearning and fine-tuning. The term $\inf_{\vartheta}\mathcal{L^{\text{ft}}}(\vartheta)$ represents the infimum (the greatest lower bound) of the fine-tuning loss function. This gives the lowest possible value of the fine-tuning loss across all parameter values, effectively acting as a lower bound for the fine-tuning loss during optimization. In this case, the constraint enforces that the fine-tuning loss of the unlearned model must not exceed the lowest fine-tuning loss found over all possible configurations of the pruned model parameters.

By applying a penalty to the constraint, we arrive at the following penalized problem: 
\begin{equation}\label{bi_penalized}
 \min_{\theta_{\text{pruned}}} \mathcal{L^{\text{penalized}}}(\theta_{\text{pruned}}),
 \vspace{-5pt}
\end{equation}
where
\begin{equation}
\begin{split}
    &\mathcal{L^{\text{penalized}}}(\theta_{\text{pruned}}):=\\
    &\mathbb{E}_{x_0,\epsilon,t,c,c'}  \| \epsilon_{\theta}(x_t, t,c') - \epsilon_{\theta_{\text{pruned}}}(x_t, t,c) \|^2 \\
    &+ \lambda \left(\mathcal{L^{\text{ft}}}(\theta_{\text{pruned}}) - \inf_{\vartheta}\mathcal{L^{\text{ft}}}(\vartheta)\right)
\end{split}
\end{equation}
with $\lambda > 0$. As $\lambda$ increases, the solution to the penalized problem approaches the solution to \cref{eq:ineq_constrained}, and thus also to \cref{eq:constrained} (see \cite{LuM24} Theorem 2 for an explicit relationship between the stationary points of \cref{eq:ineq_constrained} and those of \cref{eq:constrained}).

Note that the penalized problem in \cref{bi_penalized} is equivalent to the following minimax problem: 
 \begin{equation}\label{eq:minimax}
    \begin{split}
       \min_{\theta_{\text{pruned}}} \max_{\vartheta} G_{\lambda}(\theta_{\text{pruned}},\vartheta) ,
    \end{split}
\end{equation}
where 
\begin{equation}\label{eq:minimax_detailed}
\begin{split}
    &G_\lambda(\theta_{\text{pruned}},\vartheta):= \\
    &\mathbb{E}_{x_0,\epsilon,t,c,c'}  \| \epsilon_{\theta}(x_t, t,c') - \epsilon_{\theta_{\text{pruned}}}(x_t, t,c) \|^2 \\
    &+ \lambda \left(\mathcal{L^{\text{ft}}}(\theta_{\text{pruned}}) - \mathcal{L^{\text{ft}}}(\vartheta)\right).
\end{split}
\end{equation}
To solve \cref{eq:minimax}, we use a double-loop method. In each lower step, we fix $\theta_{\text{pruned}}$ and solve the maximization problem $ \max_{\vartheta} G_{\lambda}(\theta_{\text{pruned}}^k,\vartheta)$. Note that maximizing$G_{\lambda}(\theta_{\text{pruned}}^k,\vartheta)$ with respect to $\vartheta$ is equivalent to maximizing$-\mathcal{L^{\text{ft}}}(\vartheta)$, which in turn is equivalent to minimizing $\mathcal{L^{\text{ft}}}(\vartheta)$, \ie the fine-tuning objective. Consequently, there is no additional computational overhead compared to standard fine-tuning. In each upper step, we fix $\vartheta$ and update $\theta_{\text{pruned}}$ using the gradient of $\nabla_{\theta_{\text{pruned}}}G_{\lambda}(\theta_{\text{pruned}}^k,\vartheta)$. By comparing ~\cref{eq:concept_ablation} with ~\cref{eq:out-kd} and ~\cref{eq:out-feat-kd}, we see that both the lower and upper steps have the same computational requirements. Both involve using the diffusion model twice, with the upper level requiring slightly less computation since it does not involve the teacher base model. Thus, overall, our bilevel method has even slightly lower computational cost than a standard fine-tuning objective with distillation when the total number of iterations are equal. ~\cref{alg:bilevel} presents our bilevel algorithm.
\begin{algorithm}
\caption{Bilevel fine-tuning and concept removal for pruned diffusion models}\label{alg:bilevel}
\begin{algorithmic}[1]
\STATE{Input: Fine-tuning Data: $D_f$, target concept: $c$, anchor concept: $c'$, pruning result: $\theta_{\text{pruned}}^{0}$, Total Upper iterations: $E \in \mathbb{N}_+$, Lower iterations between two upper updates: $K \in \mathbb{N}_+$, penalty coefficient: $\lambda\ge 0$, lower and upper learning rates $\eta$ and $\zeta$. }
\FOR{$e=0,\dots,E-1$}
\FOR{$k=0,\dots,K-1$.}
\STATE{Let $\vartheta^{e,k+1}=\vartheta^{e,k} - \eta \nabla_{\vartheta} \mathcal{L}^\text{ft}(\vartheta^{e,k})$.}
\STATE{Output $\vartheta^{e,K}$.}
\ENDFOR
\STATE{Let $\theta_{\text{pruned}}^{e+1}=\theta_{\text{pruned}}^{e} - \zeta \nabla_{\theta_{\text{pruned}}} G_{\lambda}(\theta_{\text{pruned}}^{e},\vartheta^{e,K})$.}
\ENDFOR
\end{algorithmic}
\end{algorithm}
Since the gradient of $G$ with respect to  $\theta_{\text{pruned}}$ is influenced by both the upper-level and lower-level losses, this approach incorporates more information from fine-tuning in concept unlearning. This interdependency between the upper and lower levels is the key difference between our bilevel approach and a naive two-stage method.

\section{Experiments \& Results}\label{sec:exp}
\subsection{Effect of Distillation and Pruning}\label{sec:exp_sub:distill_pruning_effect}
While previous studies~\cite{BKSDM,aptp,progressive-distill-sdxl,koala,laptop-diff} have employed distillation during the fine-tuning of pruned diffusion models, to the best of our knowledge, the impact of knowledge distillation on the convergence of pruned models has not yet been quantified. We begin our experiments by examining the impact of knowledge distillation on convergence speed. We adopt APTP~\cite{aptp} as the pruning method, as it has been shown to be better-suited for T2I diffusion models than static pruning. APTP dynamically prunes a pre-trained diffusion model into a Mixture of Experts, where each expert is optimized for generating images aligned with prompts of similar complexity. A comprehensive description of APTP is provided in ~\cref{sec:app-aptp}. For our experiments, we prune Stable Diffusion 2.1~\cite{ldm-stable} using APTP. we select two experts with MAC budgets of $0.55$ and $0.8$. Although we use APTP, our proposed method is independent of the pruning strategy and can be applied as a plug-in with any static or dynamic pruning method and objective.

\begin{figure}
    \centering
    \includegraphics[width=\linewidth]{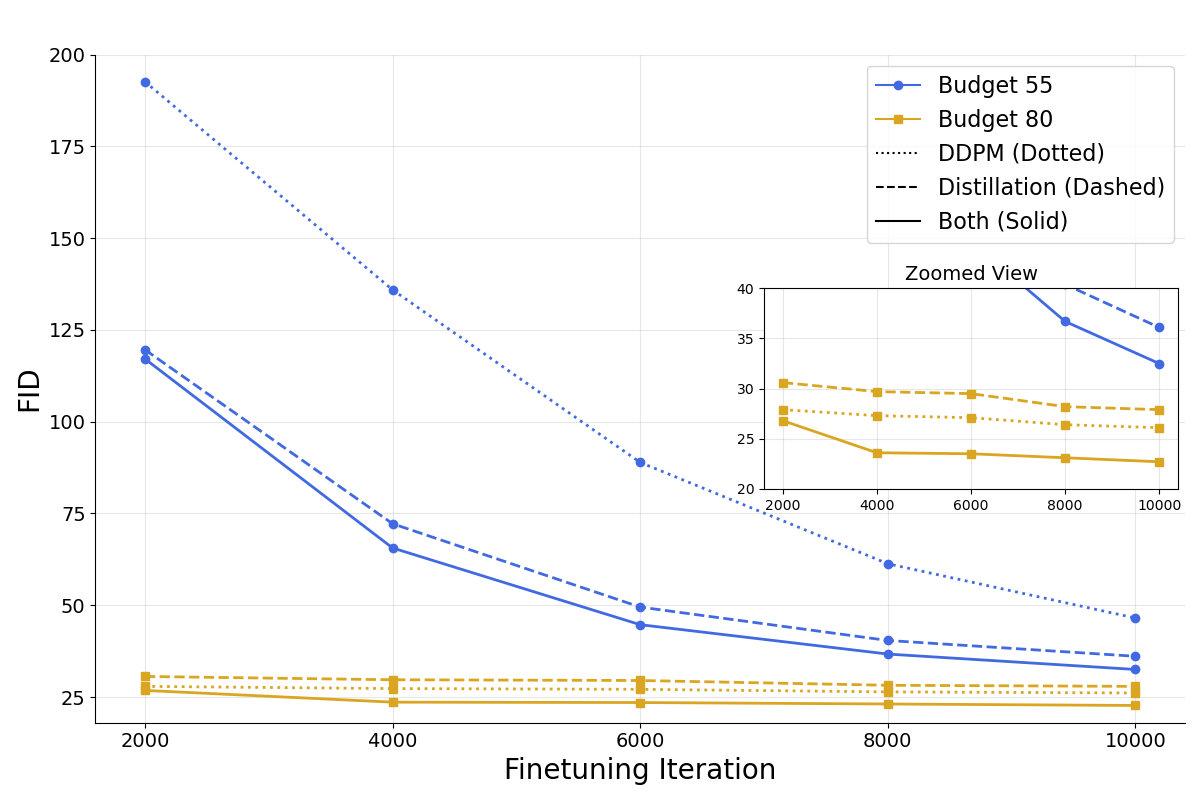}
    \caption{\textbf{Effect of Distillation:} Adding distillation to the fine-tuning process of a pruned model significantly accelerates convergence, especially in resource-constrained settings.}
    \label{fig:distill-effect}
    \vspace{-10pt}
\end{figure}

For fine-tuning, we use the MS-COCO-2017~\cite{mscoco} image-caption dataset and report FID~\cite{FID} scores on its 5,000 validation samples. \cref{fig:distill-effect} illustrates the effect of distillation on the convergence during fine-tuning of a pruned model. The results indicate that incorporating distillation into the fine-tuning objective accelerates convergence and achieves a better FID for both budget settings. Detailed Experimental settings can be found in the ~\cref{sec:app-exp_sub:details}.

\begin{figure}
    \centering
    \includegraphics[width=\linewidth]{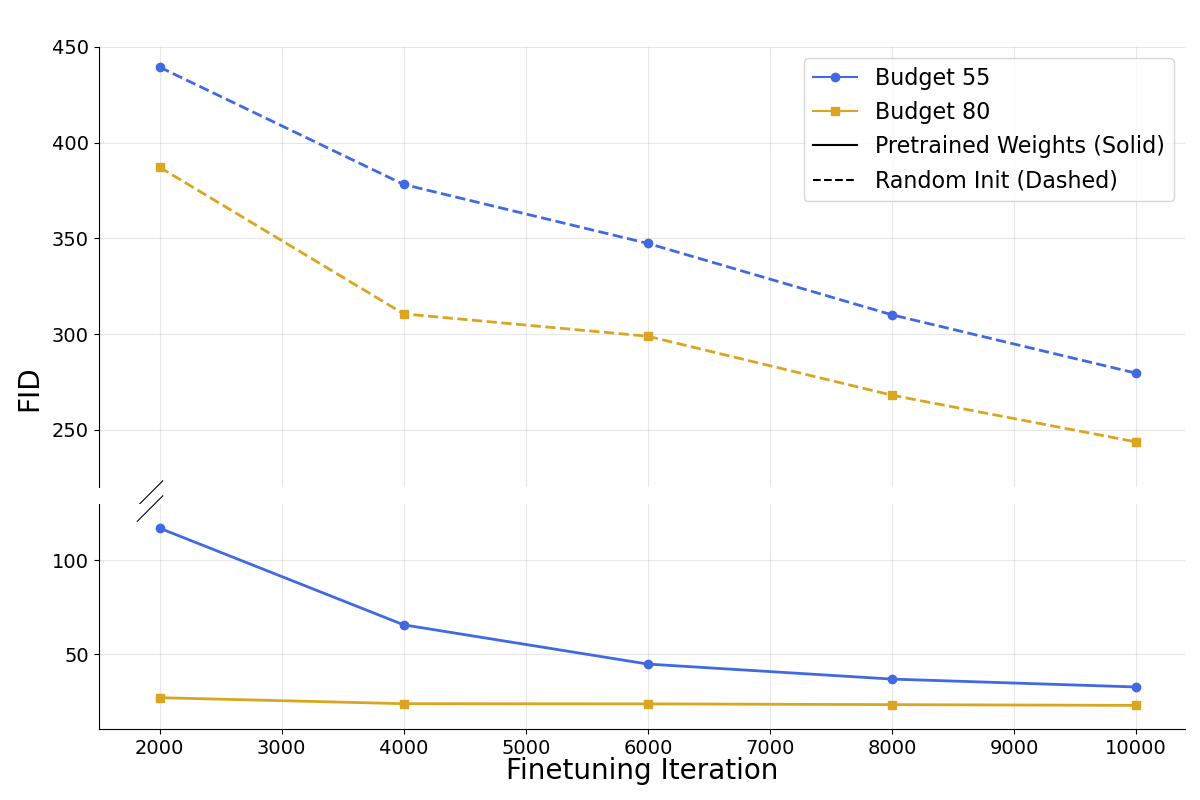}
    \caption{\textbf{Effect of Pruning:} Pruning enables significantly faster convergence compared to random initialization, making it an excellent choice for training a small diffusion model.}
    \label{fig:random-pretrained}
    \vspace{-10pt}
\end{figure}

While SPDM~\cite{spdm} demonstrates that pruning is more effective than using randomly initialized weights within the same structure, their fine-tuning stage relies solely on the diffusion loss (\cref{eq:denoising-obj}). Given the remarkable improvements provided by distillation, an important question arises: could applying distillation on a smaller, randomly initialized diffusion model yield similar benefits to pruning, thereby rendering the pre-trained weights from the base model unnecessary? In \cref{fig:random-pretrained}, we show that even with distillation, pruning provides substantial advantages over using a randomly initialized model.

\subsection{Concept Removal}
In the previous section we saw that pruning combined with distillation is effective, we now address a critical question: what if the base model contains undesirable properties that need to be removed? One option is to fine-tune the pruned model and subsequently apply an existing editing or unlearning method to eliminate unwanted concepts. As hypothesized in \cref{sec:method}, this approach may be suboptimal, and in the following sections we show quantitatively that our proposed bilevel method is more effective in this scenario.

\subsubsection{Experimental Setting}
To validate that our bilevel approach is superior to a two-stage pipeline for controlled distillation, we follow a similar setup to \cref{sec:exp_sub:distill_pruning_effect}. First, we prune the Stable Diffusion 2.1 model to an $80\%$ MAC budget on MS-COCO-2017~\cite{mscoco} training data using APTP\cite{aptp} and then fine-tune it with both denoising loss and output- and feature-level distillation (\cref{eq:ft_all}) for 20,000 iterations also using MS-COCO-2017, yielding a smaller but hiqh quality model. As shown in \cref{fig:teaser}, this fine-tuned model retains the capability to generate high quality images of various styles and concepts comparable to the original Stable Diffusion model.

Next, to complete a two-stage pipeline, we apply ESD~\cite{erasing-dm}, UCE~\cite{uce}, and ConceptPrune~\cite{concept-prune} as baselines on top of the fine-tuned model to remove some concepts from the fine-tuned model. We use the optimal hyperparameters from respective papers, detailed in the ~\cref{sec:app-exp_sub:details}. 

Additionally, we apply our bilevel framework to fine-tune the same $80\%$-MAC model and remove the same concepts. In the lower-level optimization, we perform standard fine-tuning as outlined in ~\cref{eq:ft_all}. For the upper-level optimization, any existing concept unlearning method can be used; we choose an approach similar to ESD~\cite{erasing-dm}, which applies a negative guidance step~\cite{cfg} to steer the diffusion model away from generating samples of a target concept. To ensure a fair comparison—and even to favor the baselines—we run our bilevel fine-tuning for a total of 20,000 iterations, covering both lower-level and upper-level steps. We do 20 lower steps between two upper steps. We set $\lambda$ in ~\cref{eq:minimax_detailed} to $100$. See ~\cref{sec:app-exp_sub:details} for more details.

\subsubsection{Style Removal Results}
\begin{figure}
    \centering
    \includegraphics[width=\linewidth]{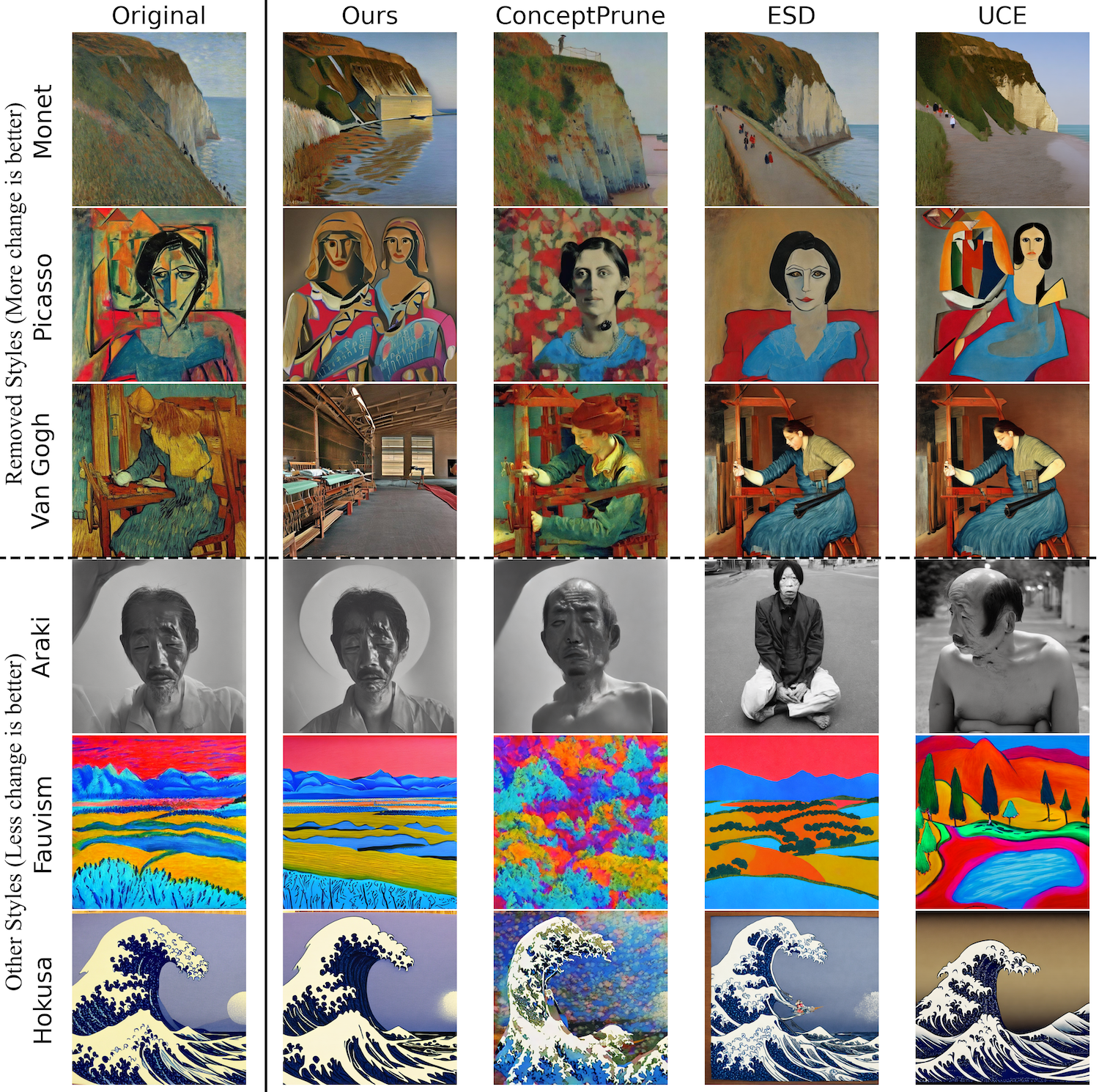}
    \caption{Quantitative results demonstrate the effectiveness of removing the styles of three artists—Monet, Picasso, and Van Gogh—from the pruned model. Our method not only successfully eliminates the target styles completely but also generates other non-removed styles more effectively than the baselines. Original refers to the pruned model that is fine-tuned using only ~\cref{eq:ft_all}.}
    \label{fig:artist_erasure}
    \vspace{-10pt}
\end{figure}

\begin{table}
    \center
    \resizebox{\linewidth}{!}{
    \begin{tabular}{lccccc}
        \toprule
        & \multicolumn{3}{c}{Artist Erasure} & \multicolumn{2}{c}{COCO} \\
        \cmidrule(r){2-4} \cmidrule(l){5-6}
        & CLIP~($\downarrow$) & CP~($\uparrow$) & CSD~($\downarrow$) & FID~($\downarrow$) & CLIP~($\uparrow$) \\
        \midrule
        Stable Diffusion 2.1~\cite{ldm-stable} & 34.44 & 44.0 & 87.91 & 15.11 & 31.60 \\ 
        Distilled Model(~\cref{eq:ft_all}) & 34.34 & 0.0 & 100.0 & 22.19 & 29.44 \\ \midrule
        Distilled Model + ESD~\cite{erasing-dm} & 30.78 & 84.0 & 61.45 & 30.38 & 29.02\\
        Distilled Model + UCE~\cite{uce} & 30.48 & 82.66 & 65.09 & 26.63 & \textbf{29.28}\\
        Distilled Model + CP~\cite{concept-prune} & 29.96 & 91.3 & 53.19 & 27.86 & 28.94 \\
        Bilevel (\textbf{Ours}) & \textbf{26.28} & \textbf{97.6} & \textbf{39.04} &\textbf{22.24} & 29.19 \\
        \bottomrule
    \end{tabular}
    }
    \caption{ \textbf{Style Removal}: 
    Quantitative results for removing the styles of three artists—Monet, Picasso, and Van Gogh—from the pruned model across 50 prompts for each artist. 
    CP Score ~\cite{concept-prune} penalizes an unlearning method if the model produces images that have a higher clip score to the style prompt than the original model. CSD~\cite{csd} is a metric specifically designed to measure style similarity. The COCO values demonstrate the model's ability to retain styles and concepts that were not targeted for removal. Our bilevel method effectively removes the target concepts while restoring the generation capabilities of the pruned model.}
    \label{tab:artist_erasure}
    \vspace{-10pt}
\end{table}

Following the evaluations from \cite{concept-prune}, we assess the effectiveness of our approach and the baselines in removing the styles of three artists—Vincent Van Gogh, Pablo Picasso, and Claude Monet—whose styles are effectively replicated by Stable Diffusion~\cite{erasing-dm}. Using a dataset of 50 prompts for each artist’s style, we report the CLIP similarity~\cite{clip} between generated samples and the prompts, as well as a stricter CLIP score that penalizes the unlearned model when its generated samples are more similar to the prompts than those of the original unpruned model~(introduced in ~\cite{concept-prune}). We also report the CSD Score~\cite{csd}, a metric specifically proposed for measuring style similarity. The standard CLIP-based metrics quantify general alignment between the generated samples and the prompts. In contrast, the CSD score specifically targets style similarity, allowing for the assessment of stylistic nuances that may be missed by CLIP-based metrics. These metrics together provide a more comprehensive evaluation by capturing different aspects of the unlearning process. 

We also evaluate our proposed method and the baseline on their retention of the generation capabilities for unrelated, unremoved concepts. We report the FID~\cite{FID} and CLIP similarity scores between generated samples and prompts on 5,000 validation samples from the MS-COCO-2017 dataset.

We present the results in  \cref{tab:artist_erasure}. Model~\cref{eq:ft_all} represents the pruned model fine-tuned using ~\cref{eq:ft_all}. We use Model~\cref{eq:ft_all} as the reference for calculating the values in \cref{tab:artist_erasure}, which explains the $0.0$ artist erasure score in the second row. We can see that the fine-tuned model is highly capable of generating various concepts and styles (Note the high artist similarity score for Model~\cref{eq:ft_all}).
Although we use ESD~\cite{erasing-dm} as the concept unlearning method in the upper step, our proposed method significantly outperforms the two-stage Distillation+ESD approach, achieving 15\% lower CLIP similarity, 16\% higher CP score, and 36\% lower CSD score. Additionally, it delivers better generation quality (lower FID) and higher CLIP scores on other unremoved concepts. Our method also outperforms other two-stage baselines with different erasing methods, particularly in terms of CSD score, which focuses on style similarity. We achieve a 27\% lower CSD score compared to the best baseline. Since our approach can incorporate different unlearning methods, we believe that leveraging more powerful removal techniques could further enhance its performance.

In ~\cref{fig:artist_erasure}, we present qualitative results comparing the concept erasure capabilities of our proposed method and the baselines. Our bilevel method effectively removes the desired concept entirely. Note the subtle similarities between the generated images of the baselines and the original model in the first three rows, where some leakage indicates that the model still retains aspects of the removed style. For instance, the prompt used for the samples in the style of Van Gogh is ``The Weaver by Vincent van Gogh". Although the original Van Gogh painting depicts a woman, the prompt does not specify the subject’s gender. However, baseline methods continue to generate an image of a woman, suggesting that residual elements of Van Gogh's painting persist in them even after removal. In contrast, our method generates high-quality images without any trace of the erased styles. Moreover, our method excels at preserving other non-removed styles, as evidenced by the comparison between images from the original model and our method in the last three rows. The baselines, however, exhibit style interference, altering other non-removed styles as well.

\subsubsection{Explicit Content Removal Results}
\begin{figure}
    \centering
    \includegraphics[width=0.9\linewidth]{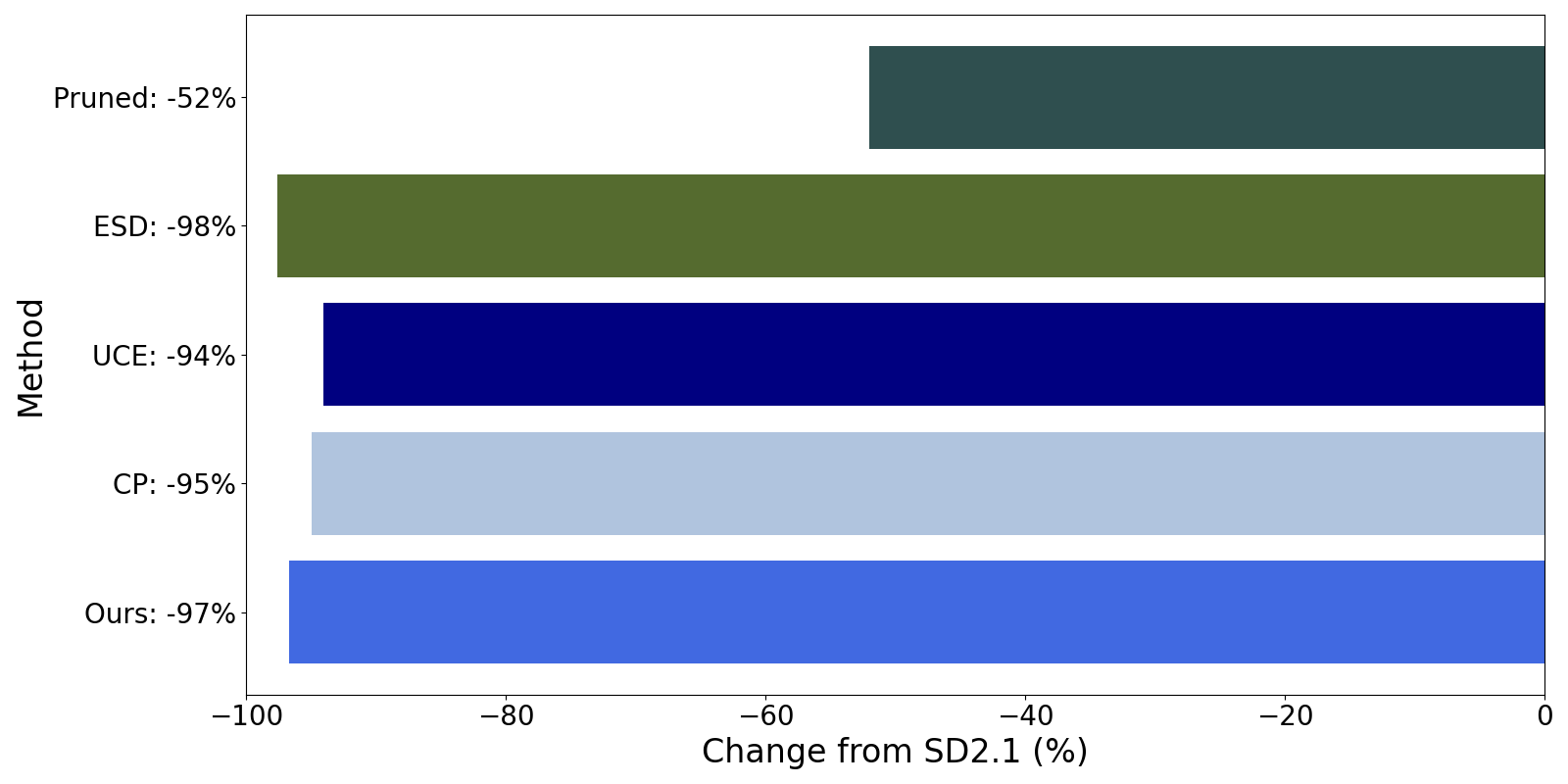}
    \caption{Explicit Content Removal: The values represent the percentage decrease in nudity content in I2P prompts compared to the original SD2.1 model. Pruned baseline performs well as seen by prior work. Our method achieves performance on par with baseline models for NSFW content reduction.}
    \label{fig:i2p_nudity_bar}
    \vspace{-10pt}
\end{figure}

We also evaluate our method's ability to remove explicit Not Safe For Work (NSFW) content. Following previous work~\cite{erasing-dm,concept-prune}, we use 4,703 prompts from the Inappropriate Prompts Dataset (I2P)\cite{i2p}. Using these prompts, we generate samples containing various inappropriate elements with our method and the baselines. The images are then classified using the Nudenet detector\cite{nudenet}. In \cref{fig:i2p_nudity_bar}, we compare the effectiveness of our approach and the baselines in reducing NSFW generation from Stable Diffusion 2.1. First, we observe that fine-tuned pruned model reduces explicit content independently, without additional measures. Additionally, our bilevel method performs as well as the baseline models in explicit content removal.

Following~\cite{concept-prune}, we evaluate the resilience of our approach on the adversarial NSFW prompt datasets, Ring-A-Bell~\cite{ring-a-bell} and MMA~\cite{mma}. Using adversarial prompts on the baseline model fine-tuned as described in \cref{eq:ft_all}, we compare the number of explicit images generated by our method and the concept removal baseline relative to the fine-tuned baseline.  \cref{tab:nsfw_erasure} shows the results. Our method shows solid performance on adversarial prompts.

\begin{table}
    \center
    \resizebox{\linewidth}{!}{
    \begin{tabular}{lccccc}
        \toprule
        Method & \multicolumn{2}{c}{NSFW Removal Bench.} & \multicolumn{2}{c}{COCO}\\
        \cmidrule(r){2-3} \cmidrule(r){4-5}
        & MMA~($\uparrow$) & Ring-a-Bell~($\uparrow$) & FID~($\downarrow$) & CLIP~($\uparrow$) \\
        \midrule
        Distilled Model + ESD~\cite{erasing-dm}               & 93.70 & 77.27  & 32.47 & 28.57 \\
        Distilled Model + UCE~\cite{uce}                      & 88.57 & 76.14 & 41.55 & 26.60 \\
        Distilled Model + CP~\cite{concept-prune}   & \textbf{94.12} & \textbf{97.72} & 29.56 & 29.45 \\ \midrule
        \textbf{Ours (ESD)}                       & 91.60 & 94.32 & \textbf{26.80} & \textbf{29.94} \\
        \bottomrule
    \end{tabular}
    }
    \caption{Comparison of our method with two-stage baselines. Our method shows competitive NSFW removal while achieving superior generation quality on concepts not targeted for removal.}
    \label{tab:nsfw_erasure}
    \vspace{-10pt}
\end{table}

When interpreting the results in \cref{fig:i2p_nudity_bar} and \cref{tab:nsfw_erasure}, it is important to consider generation quality on other unaffected concepts. As shown in \cref{tab:nsfw_erasure}. Our method performs comparably to the baselines on NSFW removal, with significantly less impact on generation quality judged by FID and CLIP scores on 5000 validation prompts of MS-COCO. Our method offers a favorable trade-off between explicit content reduction and output quality.

As mentioned before, our framework is compatible with any unlearning technique. Our baseline is a \textbf{\emph{two stage} distillation + unlearning framework} rather than a specific unlearning technique. The objective of our experiments was to demonstrate how our bilevel method even when paired with a basic unlearning approach like ESD~\cite{erasing-dm} can outperform two stage baselines with powerful unlearning methods. Paired with a more powerful unlearning method, our approach can outperform the baselines~(See~\cref{sec:app-results}).

\subsection{Continual Unlearning}
Our method produces an efficient, and unlearned checkpoint that retains its original generation capability. Any unlearning approach can be applied on the result if unlearning of another concept is needed. Nonetheless, the reasoning outlined in~\cref{sec:method} also applies to continual unlearning—i.e., a two-stage process of first unlearning $c_1$ and then unlearning $c_2$ may be suboptimal. However, we believe the degree of suboptimality would be less severe than the challenge addressed in this paper—\ie restoring the generative capability of a pruned model while suppressing unwanted concepts. This presents a compelling direction for future research.
\section{Conclusion}\label{sec:conclusion}
In this paper, we introduce a bilevel optimization framework for fine-tuning pruned diffusion models, addressing the dual challenge of restoring generative quality while effectively suppressing unwanted concepts. Our experiments show that our framework outperforms two-stage baseline methods in both generative quality and concept removal tasks, including artist style erasure and NSFW content suppression.
Our proposed method is compatible with various pruning and concept unlearning techniques. 
This method is especially valuable for deploying generative models in settings where ethical and practical constraints demand careful control over the content.
{
    \small
    \bibliographystyle{ieeenat_fullname}
    \bibliography{main}
}

\appendix
\clearpage
\setcounter{page}{1}
\maketitlesupplementary


\section{Details of APTP}\label{sec:app-aptp}

Adaptive Prompt-Tailored Pruning (APTP)~\cite{aptp} is a novel prompt-based pruning method designed for Text-to-Image (T2I) diffusion models. T2I diffusion models are computationally intensive, especially during the sampling process, making their deployment on resource-constrained devices or for large user bases challenging. APTP aims to reduce this computational cost by tailoring the model architecture to the complexity of the input text prompt.

Instead of using a single pruned model for all inputs, APTP prunes a pretrained T2I model (e.g., Stable Diffusion) into a mixture of efficient experts, where each expert specializes in generating images for a specific group of prompts with similar complexities. This is illustrated in Figure 1 of their paper.

At the heart of APTP lies a prompt router module. This module learns to determine the required capacity for an input text prompt and routes it to an appropriate expert, given a total desired compute budget. Each expert corresponds to a unique architecture code that defines its structure as a sub-network of the original T2I model. The number of experts (and corresponding architecture codes) is a hyperparameter.

The prompt router consists of three key components:
\begin{enumerate}
    \item Prompt Encoder: Encodes input prompts into semantically meaningful embeddings using a pretrained frozen Sentence Transformer model.
    \item Architecture Predictor: Transforms the encoded prompt embeddings into architecture embeddings, bridging the gap between prompt semantics and the required architectural configuration.

    \item Router Module: Maps the architecture embeddings to specific architecture codes. To prevent all codes from collapsing into a single one, the router module employs optimal transport during the pruning phase. The optimal transport problem aims to find an assignment matrix Q that maximizes the similarity between architecture embeddings and their assigned architecture codes while ensuring equal distribution of prompts to different experts. This optimal assignment matrix is calculated using the Sinkhorn-Knopp algorithm and is used to route architecture embeddings to architecture codes during pruning.

\end{enumerate}

The prompt router and architecture codes are trained jointly in an end-to-end manner using a contrastive learning objective.

\section{Method for solving the bilevel problem}
\label{supp_method}
Classical methods for solving a bilevel problems such as \cref{eq:constrained} require calculating second order information, please see ~\citep{FrDoFrPo17, FrFrSaGrPo18, GhWa18} For examples. However, when fine-tuning foundation models, this process becomes extremely expensive due to the high computational and memory demands.
Recently, new frameworks for bilevel optimization have been introduced~\cite{LuM24, ShenC23, LiuLYZZ24, KwonKWN23, LiuYWSL22}. These methods only use first-order information and thus significantly reduce computational costs, making them extremely suitable for fine-tuning foundation models. We employ this type of method for solving \cref{eq:constrained}.

More, specifically, \cref{eq:constrained} is equivalent to the following constrained optimization problem:
\begin{equation}\label{eq-sup:ineq_constrained}
\begin{split}
    \min_{\theta_{pruned}}&\mathbb{E}_{x_0,\epsilon,t,c,c'}  \| \epsilon_{\theta}(x_t, t,c') - \epsilon_{\theta_{pruned}}(x_t, t,c) \|^2,\\
     s.t.\ & L^{ft}(\theta_{pruned}) - \inf_{\vartheta}L^{ft}(\vartheta)\le 0.
\end{split}
\end{equation}
By penalizing the constraint, we obtain the following penalized problem:
\begin{equation}\label{sup:bi_penalized}
 \min_{\theta_{pruned}} L_{penalized}(\theta_{pruned}),
\end{equation}
where
\begin{equation}
    \begin{split}
&L_{penalized}(\theta_{pruned}):=\\
&\mathbb{E}_{x_0,\epsilon,t,c,c'}  \| \epsilon_{\theta}(x_t, t,c') - \epsilon_{\theta_{pruned}}(x_t, t,c) \|^2 \\
&+ \lambda \left(L^{ft}(\theta_{pruned}) - \inf_{\vartheta}L^{ft}(\vartheta)\right)
    \end{split}
\end{equation}
and $\lambda>0$. As $\lambda$ increases, the solution to the penalized problem approaches the solution to \cref{eq-sup:ineq_constrained}, and thus the solution to \cref{eq:constrained} (see \cite{LuM24} Theorem 2 for an explicit relationship between the stationary points of \cref{eq-sup:ineq_constrained} and those of the original problem \cref{eq:constrained}).
 Note that the penalized problem \cref{bi_penalized} is equivalent to the following minimax problem:
\begin{equation}\label{minimax}
    \begin{split}
       \min_{\theta_{pruned}} \max_{\vartheta} G_{\lambda}(\theta_{pruned},\vartheta) ,
    \end{split}
\end{equation}
where 
\begin{equation}
    \begin{split}
&G_\lambda(\theta_{pruned},\vartheta):= \\
&\mathbb{E}_{x_0,\epsilon,t,c,c'}  \| \epsilon_{\theta}(x_t, t,c') - \epsilon_{\theta_{pruned}}(x_t, t,c) \|^2 \\
&+ \lambda \left(L^{ft}(\theta_{pruned}) - L^{ft}(\vartheta)\right).
    \end{split}
\end{equation}
To solve \cref{minimax}, we use a double loop method. At step $t$, we fix  $\theta_{pruned}^t$ and then solve the maximization problem $ \max_{\vartheta} G_{\lambda}(\theta_{pruned}^t,\vartheta)$. Then we update $\theta_{pruned}$ using the gradient of $\nabla_{\theta_{pruned}}G_{\lambda}(\theta_{pruned}^t,\vartheta)$. Since the gradient of $G$ with respect to $\theta_{pruned}$ is determined both by the upper loss and lower loss, this incorporates more information from feature distillation when doing concept unlearning. Therefore, the upper and lower level problems are dependent on each other. This is the key difference between the two-stage method and our bilevel method.

\section{Experiments}\label{sec:app-exp}

\subsection{Detailed experimental setup}\label{sec:app-exp_sub:details}
\subsubsection{Datasets}
In all our experiments, we use the MS-COCO Captions 2017~\cite{mscoco} with approximately 500k training image-caption pairs. For evaluations, we use the validation data of MS-COCO-2017 with 5000 images. We sample one caption per image from the validation set.

\subsubsection{Effect of Distillation and Pruning Experimental Setting}\label{sec:exp-distill}
We utilize one of the pre-trained APTP~\cite{aptp} experts on COCO, which achieves $80\%$ MAC utilization compared to the original Stable Diffusion 2.1 model~\cite{ldm-stable}. The model is fine-tuned using various objectives at a fixed resolution of $512 \times 512$ for all configurations. Optimization is performed with the AdamW~\cite{adamw} optimizer, using parameters $\beta_1 = 0.9$ and $\beta_2 = 0.999$, no regularization, and a constant learning rate of $10^{-6}$, coupled with a 250-iteration linear warm-up. Fine-tuning is conducted with an effective batch size of 64, distributed across 8 NVIDIA A6000Ada 48GB GPUs, each with a local batch size of 8.

In experiments combining DDPM and distillation losses, we compute a weighted average of the loss terms as follows:
\begin{itemize}
    \item \textbf{Diffusion loss}: weight = $1.0$
    \item \textbf{Distillation loss}: weight = $2.0$
    \item \textbf{Feature distillation loss}: weight = $0.1$
\end{itemize}

For sample generation, we employ classifier-free guidance~\cite{cfg} with a guidance scale of $7.5$ and 25 steps of the PNDM sampler~\cite{liu202pndm}. We calculate FID~\cite{FID} on the validation set of COCO-2017 for \cref{fig:distill-effect,fig:random-pretrained}.

\subsubsection{Concept Removal Experimental Settings}
In a two-stage pipeline, we first fine-tune the expert described in \cref{sec:exp-distill} for 20,000 iterations using DDPM, incorporating both output and feature distillation objectives. The fine-tuning settings are identical to those detailed in \cref{sec:exp-distill}.

\paragraph{Baselines}
We use ESD~\cite{erasing-dm}, UCE~\cite{uce} and ConceptPrune~\cite{concept-prune} as the concept removal methods for a two-stage distillation-then-forget pipeline. Details of each method follows:

\begin{itemize}
    \item 
    \textbf{ESD~\cite{erasing-dm}:} ESD is a method for erasing concepts from text-to-image diffusion models by fine-tuning the model weights using negative guidance. The goal is to reduce the probability of generating images associated with a specific concept, represented by $P_{\theta}(x) \propto \frac{P_{\theta^*}(x)}{P_{\theta^*}(c|x)^\eta}$, where $P_{\theta}(x)$ is the distribution of the edited model, $P_{\theta^*}(x)$ is the distribution of the original model, $c$ is the concept to be erased, and $\eta$ is a scaling factor. By manipulating the gradient of the log probability, the authors arrive at a modified score function: $\epsilon_{\theta}(x_t, c, t) \leftarrow \epsilon_{\theta^*}(x_t, t)- \eta[\epsilon_{\theta^*}(x_t, c, t) - \epsilon_{\theta^*}(x_t, t)]$. This function guides the model away from the undesired concept during fine-tuning. The method uses the model's existing knowledge of the concept to generate training samples, eliminating the need for additional data. ESD offers two variations: ESD-x for prompt-specific erasure, such as artistic styles, and ESD for global erasure, such as nudity. Similar to the original paper we remove "Van Gogh", "Claude Monet", and "Picasso" from the diffusion model for artist erasure, and remove "nudity" for explicit content erasure. This process uses the AdamW~\cite{adamw} optimizer with a learning rate of $0.00001$, and a negative guidance $\eta = 1$. The model is trained for 1000 iterations to remove the concept. For artist style and explicit content removal we pick "ESD-x" and "ESD-u", respectively.
    \item 
    \textbf{UCE~\cite{uce}:} UCE is a method for editing multiple concepts in text-to-image diffusion models without retraining. UCE works by directly modifying the attention weights of the model in a closed-form solution, making it efficient and scalable. The method aims to address various safety issues such as bias, copyright infringement, and offensive content, which previous methods have tackled separately. UCE modifies the cross-attention weights, denoted as $W$, to minimize the difference between the model's output for the concepts to edit, $c_i$, and their desired target output, $v^{*}_i$. This is achieved by minimizing the objective function: $\sum_{c_i in E}||Wc_i - v^{*}_i ||^2_2 + \sum_{c_j \in P}||Wc_j - W^{old}c_j||^2_2$ where E represents the set of concepts to edit and P represents the set of concepts to preserve. This formula ensures that the model's output for the edited concepts is steered towards the desired target, while preserving the output for the concepts that should remain unchanged. Identical to the original setting of the paper, we remove "Van Gogh", "Claude Monet", and "Pablo Picasso" for artist erasure and guide them towards "art". We remove "nudity" for explicit content removal and guide them towards "person". Other hyperparameters are identical to the values set in their training code.
    
    \item \textbf{ConceptPrune~\cite{concept-prune}} ConceptPrune is a method for removing unwanted concepts from pre-trained text-to-image diffusion models without any retraining. This is achieved by pruning or zeroing out specific neurons within the model's feed-forward networks that are identified as being responsible for generating the unwanted concept. This method is inspired by the observation that certain neurons in neural networks specialize in specific concepts. ConceptPrune first Identifies skilled neurons by analyzing the activation patterns of neurons in response to prompts with and without the unwanted concept and then prunes them. For ConceptPrune, we set skill ratio to $0.01$. We remove "Van Gogh", "Claude Monet", and "Picasso" from the diffusion model for artist erasure, and remove "nudity" for explicit content erasure. Other hyperparameters are identical to best settings in their released code.
\end{itemize}

\paragraph{Bilevel Experimental Setting}
For fine-tuning the pruned model according to our bilevel training setting, we use the same hyperparameters as the standard fine-tuning objective mentioned in \cref{sec:app-exp}. We do 20 lower steps between two upper steps. We set $\lambda$ in ~\cref{eq:minimax_detailed} to $100$. In each upper level step we do a step identical an ESD~\cite{erasing-dm} step. Each lower level step in our approach is identical to a standard fine-tuning with denoising and distillation mentioned for the two-stage method. We set the upper learning rate to $5e-6$.

\subsection{More Results}\label{sec:app-results}
Our framework is compatible with any unlearning technique. Our baseline is a \textbf{\emph{two stage} distillation + unlearning framework} rather than a specific unlearning technique. The objective of our experiments was to demonstrate how our bilevel method even when paired with a basic unlearning approach like ESD can outperform two stage baselines with powerful unlearning methods.

\subsubsection{Unlearn - Prune - Distill Baseline}
A natural question would be to compare our method with an alternative baseline where the concepts are first unlearnt from base model and then distillation is done. This baseline requires more resources than our method since unlearning occurs on the unpruned model. However, our analysis applies to this approach too. We conduct an experiment for this baseline (See~\cref{tab:new-results} rows 1 and 2), and the generation quality is worse. This is expected—applying unlearning to the base model reduces its quality. Using this degraded model for distillation further impacts the performance of the already weaker pruned model.

\begin{table}
    \centering
    \resizebox{0.9\linewidth}{!}{
    \begin{tabular}{lcc}
        \toprule
        Method & ASR~($\downarrow$) & FID~($\downarrow$) \\
        \midrule
        Distilled Model + ESD~\cite{erasing-dm}              & 62.7 & 32.47 \\
        ESD~\cite{erasing-dm} + Distilled Model              & 59.8 & 39.11 \\ \hline
        Distilled Model + UCE~\cite{uce}                       & 67.6 & 41.55 \\
        Distilled Model + CP~\cite{concept-prune}    & 54.9 & 29.56 \\
        Distilled Model + AdvUnlearn~\cite{zhang2024defensive-adv-unlearning}   & 36.6 & 36.17 \\ \hline
        \textbf{Ours (ESD~\cite{erasing-dm})}                       & 57.0 & \textbf{26.80} \\
        \textbf{Ours (AdvUnlearn~\cite{zhang2024defensive-adv-unlearning})}                       & \textbf{32.4} & \textbf{26.91} \\
        \bottomrule
    \end{tabular}
    }
    \caption{Attack Success Rate of adversarial NSFW prompts from ~\cite{genererate-or-not-short} and generation quality on COCO-Val-2017}
    \label{tab:new-results}
\end{table}

To show this compatibility with other unlearning methods, we ran two additional experiments: (1) Two-stage pipeline with the recently proposed AdvUnlearn~\cite{zhang2024defensive-adv-unlearning} and (2) our bilevel framework with AdvUnlearn~\cite{zhang2024defensive-adv-unlearning}.  Results in~\cref{tab:new-results}  confirm our method integrates well with AdvUnlearn~\cite{zhang2024defensive-adv-unlearning} and achieves superior performance in terms of ASR and FID compared to all baselines.

\begin{figure*}[t]
  \centering
  \includegraphics[width=\linewidth]{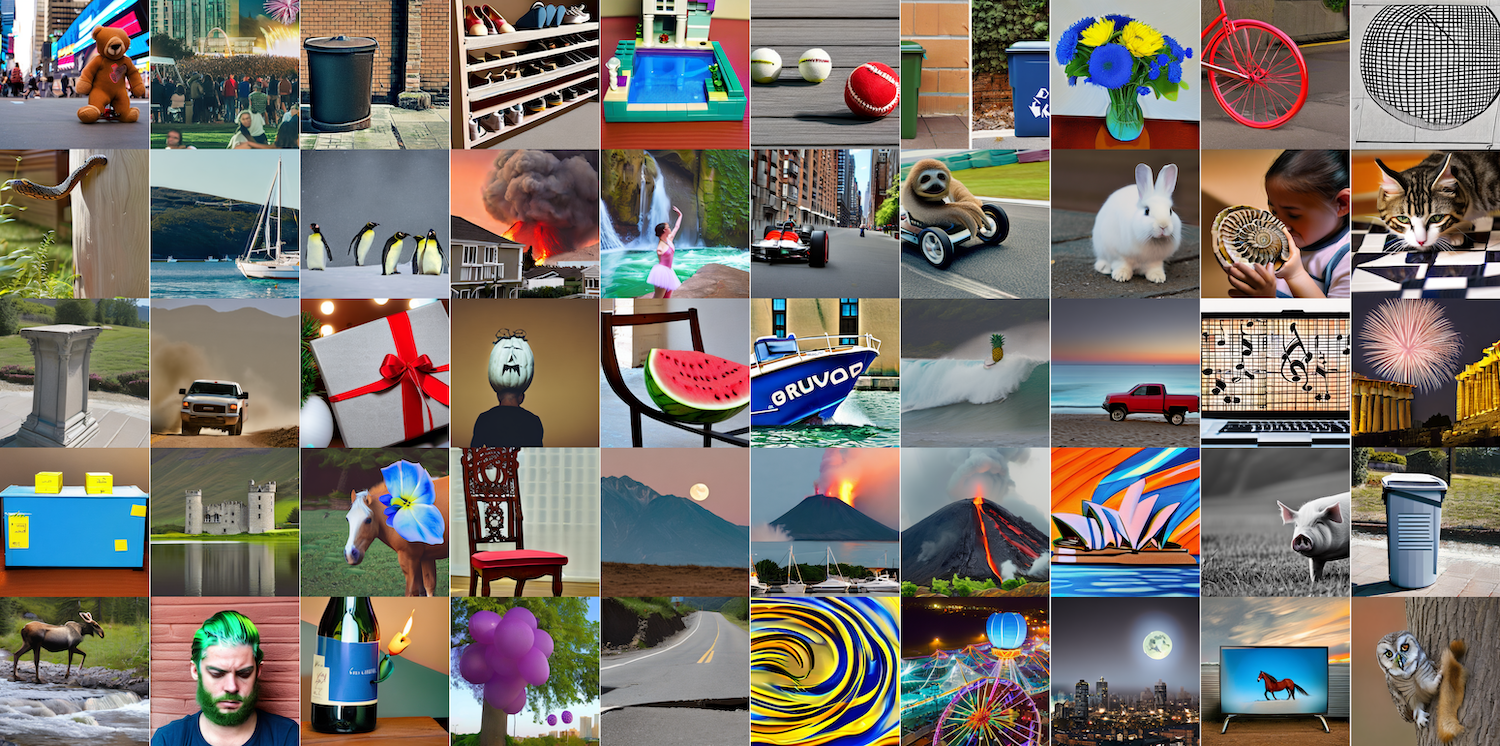}
   \caption{More visual samples of our bilevel method(with ESD)}
   \label{fig:more_samples}
\end{figure*}

We  provided more samples from our method in~\cref{fig:more_samples}.

\subsection{Figure Prompts}\label{sec:app-exp_sub:prompts}
Samples in \cref{fig:main-teaser} are generated by the prompts in \cref{tab:main-teaser-prompts}. The prompts used for \cref{fig:teaser} are presented in \cref{tab:teaser-prompts}. \cref{tab:artist-erasure-prompts} shows the prompts for generating the samples inf \cref{fig:artist_erasure}.

\begin{table}[ht]
    \centering
    \resizebox{\linewidth}{!}{
        \begin{tabular}{@{}l@{}}
        \toprule
        \textbf{Prompts} \\
        \midrule
        The Artist's House at Argenteuil by Claude Monet \\
        Child in a Straw Hat by Mary Cassatt\\
        
        \bottomrule
    \end{tabular}
}
    \caption{Prompts for Fig. \ref{fig:main-teaser}}
    \label{tab:main-teaser-prompts}
\end{table}

\begin{table}[ht]
    \centering
    \resizebox{\linewidth}{!}{
        \begin{tabular}{@{}l@{}}
        \toprule
        \textbf{Prompts} \\
        \midrule
        Water Lilies by Claude Monet \\
        The Three Dancers by Pablo Picass\\
        Red Vineyards at Arles by Vincent van Gogh \\
        A landscape with bold, unnatural colors fauvism style\\
        Girl with a Pearl Earring by Johannes Vermeer \\
        Night In Venice by Leonid Afremov\\
        The Great Wave of Kanagawa by Hokusai\\
        A watercolor painting of a forest \\
        \bottomrule
    \end{tabular}
}
    \caption{Prompts for Fig. \ref{fig:teaser}}
    \label{tab:teaser-prompts}
\end{table}

\begin{table}[ht]
    \centering
    \resizebox{\linewidth}{!}{
        \begin{tabular}{@{}l@{}}
        \toprule
        \textbf{Prompts} \\
        \midrule
        The Cliff Walk at Pourville by Claude Monet \\
        Portrait of Dora Maar by Pablo Picasso\\
        The Weaver by Vincent van Gogh \\
        Photo of a sad man by Nobuyoshi Araki\\
        A landscape with bold, unnatural colors fauvism style \\
        The Great Wave of Kanagawa by Hokusa\\
        \bottomrule
    \end{tabular}
}
    \caption{Prompts for Fig. \ref{fig:artist_erasure}}
    \label{tab:artist-erasure-prompts}
\end{table}

\end{document}